\documentclass[12pt]{article}
\usepackage{xcolor}
\usepackage{tcolorbox}
\usepackage{amsmath}
\usepackage{graphicx,psfrag,epsf}
\usepackage{enumerate}
\usepackage{natbib}
\usepackage[colorlinks=true, linkcolor=black, citecolor=blue, filecolor=black, urlcolor=black]{hyperref}
  
\tcbuselibrary{breakable}
\newcommand{\blind}{0}

\begin{document}

\def\spacingset#1{\renewcommand{\baselinestretch}%
{#1}\small\normalsize} \spacingset{1}


\if0\blind
{
  \title{\bf Searching for Best Practices in Medical Transcription with Large Language Model}
  \author{Jiafeng Li\thanks{
    Corresponding author, jl3246@cornell.edu}\hspace{.2cm}\\
    Galileo Financial Technologies, Utah, USA\\
    and \\
    Yanda Mu\thanks{
    ym379@cornell.edu}\hspace{.2cm} \\
    Lazard Inc, New York, USA}
  \maketitle
} \fi

\if1\blind
{
  \bigskip
  \bigskip
  \bigskip
  \begin{center}
    {\LARGE\bf Title}
\end{center}
  \medskip
} \fi

\bigskip
\begin{abstract}
The transcription of medical monologues, especially those containing a high density of specialized terminology and delivered with a distinct accent, presents a significant challenge for existing automated systems. This paper introduces a novel approach leveraging a Large Language Model (LLM) to generate highly accurate medical transcripts from audio recordings of doctors' monologues, specifically focusing on Indian accents. Our methodology integrates advanced language modeling techniques to lower the Word Error Rate (WER) and ensure the precise recognition of critical medical terms. Through rigorous testing on a comprehensive dataset of medical recordings, our approach demonstrates substantial improvements in both overall transcription accuracy and the fidelity of key medical terminologies. These results suggest that our proposed system could significantly aid in clinical documentation processes, offering a reliable tool for healthcare providers to streamline their transcription needs while maintaining high standards of accuracy.
\end{abstract}

\spacingset{1.45}
\section{Introduction}
\label{sec:intro}

In the realm of healthcare, the accurate and timely transcription of medical records plays a pivotal role in patient care, clinical research, and healthcare administration. The need for precise medical transcription is increasingly critical as healthcare systems strive for efficiency, accuracy, and improved patient outcomes. However, existing approaches to medical transcription often fall short, particularly when confronted with the complexities of diverse accents and specialized medical terminology that characterize doctor-patient interactions.

The challenge lies not only in the technical fidelity of converting spoken medical notes into text but also in ensuring that the nuances of medical jargon and regional accents are correctly interpreted and transcribed. Doctors, with their varied linguistic backgrounds and the specialized lexicon of medical science, present a formidable barrier to conventional transcription methods.

The repercussions of inaccurate transcription are profound, impacting clinical decision-making, treatment protocols, and patient safety. A single misinterpreted term can lead to potentially grave consequences, underscoring the critical need for more robust and reliable transcription solutions in healthcare settings.

Recent advancements in Artificial Intelligence (AI) and Natural Language Processing (NLP) offer promising avenues for addressing these challenges. Large language models, such as those based on the Transformer architecture, have demonstrated remarkable capabilities in understanding and generating human-like text. These models, trained on vast datasets, can decipher complex medical terminology and accommodate a range of linguistic variations, including accents. However, despite their advancements, the performance of current AI models in medical transcription still leaves considerable room for improvement, especially in the context of accented language. Common medical dialogue datasets  \citep{chen2020meddialog} don't really provide much opportunity for training on accented language. Challenges persist in accurately capturing the context-specific nuances of medical discourse \citep{qiu2024application}.

This paper explores the current landscape of medical transcription, highlighting the limitations of traditional methods and the potential of AI-driven approaches to revolutionize the field. By leveraging recent research and practical applications of ASR and transcript correction technologies, this study aims to underscore the urgency and benefits of adopting more sophisticated tools for achieving accurate and reliable medical transcripts and patient records.

\section{Dataset}
For this research, we used a dataset provided by Dr.~Vijai Nagarajan, who's an interventional cardiologist with rich experience in the field of cardiology, and sees the urgency of better medical transcription approach for better transcribing performance, especially in the case where the original audio consists heavy accents as well as medical jargon. The dataset Dr. Nagarajan provided contains 7 recordings, each is around 3 minutes long. The recording is a monologue of a doctor, stating the patients medical records including past medical history, examinations, and assessment and plan. Each recording contains around 300 words. The speech speed is fast, and it is very hard for untrained interpreter to understand and transcribe. 5 of the 7 provided recordings have corresponding correct transcript, so we used those 5 for experimenting. The content that involves personal information such as name, medical record number, are all made up, so there's no real concern of personal data leaking.

We present the following as an example of the correct transcript of one of the recordings:

\begin{tcolorbox}[breakable,boxrule=0pt, title=Recording 1 Transcript]
80-year-old female with complex medical history including diabetes mellitus, hypertension, dyslipidemia, obesity, anxiety/depression, breast cancer in the past, atrial fibrillation on anticoagulation with apixaban, heart failure with LVEF around 40\% presented to hospital with bradycardia and underwent permanent pacemaker placement.  She has been doing well since that time without any cardiovascular symptoms. Xarelto has been resumed and patient is back on amiodarone 200 mg p.o. daily.  She denies any new issues.
Past medical history: as below 
Medications: reviewed 
Examination: Temperature 98.2, pulse 60 bpm, blood pressure 120/80, saturation 97\% General: Patient alert oriented comfortable 
CVS: Regular no murmur 
Lungs: No wheeze or crackles 
Abdomen: Soft, nontender 
Legs: Trace bilateral pedal edema 
CNS: No focal neurological deficit
Skin: Normal 
Pacemaker site: No bleeding or hematoma 
Labs: Sodium 140, potassium 3.8, BUN 12, creatinine 1.02, TSH is normal, WBC 4 hemoglobin 13, platelets 150.
Chest x-ray: pacemaker lead in the correct position. 
Echocardiogram showed LVEF of 55 to 60\% and mild tricuspid regurgitation.
Assessment and plan\\
1. Complete heart block\\
Patient presented with bradycardia - Currently on complete heart block.  Patient was not on any AV nodal blocking agents.  Patient is on amiodarone.  Amiodarone was stopped and patient was taken permanent pacemaker and replacement.  Patient is doing well without any bleeding or hematoma.  Amiodarone has been restarted.  Recommended follow-up with electrophysiology team. \\
2. Atrial fibrillation\\
History of paroxysmal atrial fibrillation. (unclear dictation) Patient was on apixaban which has been resumed.  Monitor closely for any bleeding or hematoma.\\
3. Hypertension\\
Blood pressure is well-controlled on lisinopril 5 mg p.o. daily. Home blood pressure and heart rate monitoring recommended.  
\end{tcolorbox}

\section{Methods}
\label{sec:meth}
The first step of our approach is to obtain a base transcript from a LLM powered ASR, such as open-AI whisper. Although LLM based ASRs are showing promising performance nowadays, when given an audio that mixes terminologies and accents at the same time, along with a fast speech speed, its performance doesn't seem to be that satisfying. For example, the aforementioned recording 1, after being processed by openAI-whisper model (whisper-1), gives the following output:\\ 
\begin{tcolorbox}[breakable,boxrule=0pt, title=Recording 1 Transcript]
Eight-year-old female with complex asthma-related shrinking diabetes and some high blood pressure, come under QPDMA, obesity, anxiety, depression, breast cancer in the past, atrial fibrillation on anticoagulation topics abound, heart failure with LVF around 40\%, hospital with bradycardia and underwent permanent pacemaker placement. Period, she has been doing well since the time without any cardiovascular symptoms. Period, Xeralta has been resumed and patient is back on amino 2 and will be on POO daily. Period, iodine is in new shoes. Period, past medical assurance, no medication is available. All I see is none. Examination temperature, 98.2. Pulse 60 beats per minute. Blood pressure 120 over 80. Succession, 97\%. General patient allowed to enter if comfortable. Serious, regular, normal. Lungs, no visual crackles. Abdomen soft, no tendons, legs. Trace barotrauma, CNS, no focal neurological risk. Skin normal. Pacemaker site, no bleeding or hematoma. Blast sodium 140, potassium 3.8, BUN 12, creatinine 1.02. TSH is normal. WBC 4, hemoglobin is 13, platelets is 150. Chest x-ray, pacemaker leads are in correct position. Period. X-radiogram showed LV of 55 to 60\%. And mild tricuspid regurgitation. Period, assayment plan. Problem number one, complete heart block. Next in patients with diuretic cardiac does not mean complete heart block. Period, patient was not on any immunoblocking agents. Period, patient has an amiodarone. Period, amiodarone was stopped and patient has taken for permanent pacemaker replacement. Period, patient is doing well. Without any bleeding or hematoma. Period, amiodarone has been restarted. Period, recommended follow-up with electrophysiology team. Period, problem number two, natal fibrillation. Next in patients with paroxysmal natal fibrillation. Period, patient was on a nitrone. Nitrone has been resumed. Period, patient was on apixaban. Nitrone has been resumed. Period, monitor closely for any bleeding or hematoma. Period, problem number three, hypertension, blood pressure is well controlled. Unless no proof of amyloid in a few days. Home blood pressure and heart rate monitoring recommended.
\end{tcolorbox}

Word Error Rate (WER) is a standard measure used to assess the accuracy of ASR
systems by comparing the transcribed, hypothesis text against a reference transcript.
The WER is calculated using the formula:\\
$$W=\frac{S + D + I}{N}$$
, where S is the number of substitutions, D is the number of deletions, I is the number
of insertions, and N is the number of words in the reference transcript.

The WER of recording 1 transcript is 50.2\%, which is generally considered very high. In addition to WER, we also manually pick out medical terminologies in each recording and check it's transcription error rate, because those are really the most important words we want to ensure that are transcribed correctly. Using recording 1 transcript as an example, the medical terms we picked are 
\begin{tcolorbox}[breakable,boxrule=0pt]
"mellitus, hypertension, dyslipidemia, anticoagulation, apixaban, LVEF, bradycardia, cardiovascular, Xarelto, amiodarone, p.o., saturation, Abdomen,  edema, neurological, hematoma, sodium, potassium, BUN, creatinine, TSH, WBC, hemoglobin, platelets, echocardiogram, tricuspid regurgitation, electrophysiology, paraxysmal atrial fibrillation, lisinopril"
\end{tcolorbox}
In total there are 27 terminologies, and 15 of them are correctly transcribed, giving us a 44.4\% error rate.

\subsection{ASR transcript correction by LLM in one set}

In this session, we present the result of transcription corrected by LLM using prompt engineering. This method is inspired by another recent research conducted by researchers at Google Cloud \citep{adedeji2024sound}. Specifically, we are inputting the transcription into LLM in one set. In comparision, in session 3.2, we will input the transcription sentence by sentence. We input the whole ASR transcript into LLM model with the following prompt:\\
\begin{tcolorbox}[breakable,boxrule=0pt]
Here is a text from a medical transcript obtained from an ASR. It might contain some error, for example, wrong medical term transcribed. Can you help me correct it by replacing the words that you think are most likely the wrong ones, with words that you think are most possibly the right word in this context? You can also add words or delete words. The words are mostly in the field of cardiology, so please try to relate to cardiological terms. If you think it is correct already, leave it without any change. Here is the text: \{...\}
\end{tcolorbox}
The output from LLM model is slightly improved, but the WER and KMTER are still relatively high, 62.1\% and 36.2\% respectively.

\subsection{ASR transcript sentence by sentence correction by LLM}
In this session, we present the result of transcription corrected by LLM using prompt , sentence by sentence. We first parse the original ASR generated transcript into single sentences, then using the following prompts, input each sentence and get corresponding correction and reasoning.
\begin{tcolorbox}[breakable,boxrule=0pt]
Here is a sentence from a medical transcript obtained from an ASR. It might contain some error, for example, wrong medical term transcribed. Can you help me correct it by replacing the words that you think are most likely the wrong ones, with words that you think are most possibly the right word in this context? You can also add words or delete words. The words are mostly in the field of cardiology, so please try to relate to cardiological terms. If you think it is correct already, leave it without any change. Here is the sentence: \{...\}
\end{tcolorbox}
For example, with the following sentence from recording 1:
\begin{tcolorbox}[breakable,boxrule=0pt, colback={pink!45}]
Xeralta has been resumed and patient is back on amino 2 and will be on POO daily
\end{tcolorbox}
If we insert above sentence into the prompt, we'll get following correction and reasoning:
\begin{tcolorbox}[breakable,boxrule=0pt, colback={pink!45}]

"Xarelto has been resumed, and the patient is back on amiodarone and will be on PO (per os) daily."

Explanation:

1. Xeralta should be Xarelto, which is a commonly prescribed anticoagulant.\\
2. amino 2 seems to be a mistranscription of amiodarone, an antiarrhythmic medication.\\
3. POO should be corrected to PO, which stands for "per os," meaning "by mouth" in medical terminology.
\end{tcolorbox}
All three corrections are correct in terms of wording. Referring to the correct answer we provided in the introduction of this paper, we can see that medical terminologies such as "xeralto", "amiodarone" and "p.o." are all corrected.

Here's another example to show it's great performance, sentence:
\begin{tcolorbox}[breakable,boxrule=0pt, colback={yellow!45}]
Xeralta has been resumed and patient is back on amino 2 and will be on POO daily
\end{tcolorbox}

Correction and reasoning:
\begin{tcolorbox}[breakable,boxrule=0pt, colback={yellow!45}]
"Xarelto has been resumed, and the patient is back on amiodarone and will be on PO (per os) daily."

Explanation:

1. Xeralta should be Xarelto, which is a commonly prescribed anticoagulant.\\
2. amino 2 seems to be a mistranscription of amiodarone, an antiarrhythmic medication.\\
3. POO should be corrected to PO, which stands for "per os," meaning "by mouth" in medical terminology.
\end{tcolorbox}
However, LLM models can not always make precise correction and reasonings at all time. With the following example sentence from recording 1 transcript, we can see drawbacks of this approach too:
\begin{tcolorbox}[breakable,boxrule=0pt, colback={green!15}]
 Pacemaker site, no bleeding or hematoma.
\end{tcolorbox}
This sentence from the ASR base transcript is already fully correct, however, even if we prompted the LLM GPT that if it is correct, then leave it without changing anything, the output still replaced a key medical term with a wrong one:
\begin{tcolorbox}[breakable,boxrule=0pt, colback={green!15}]
Pacemaker site, no bleeding or ecchymosis.
\end{tcolorbox}
Therefore, to minimize additional error from LLM, we propose the manual plus LLM sentence by sentence correction method, introduced in the following session.
\subsection{Manual + LLM sentence by sentence correction of ASR transcript }
From the above two sessions, we can easily see that LLM powered ASR could generate a base transcript that contains some correct medical terms, while LLM model could further correct it, but also possibly change a correct word to a wrong one. Observing this, we propose a transcription method that combines manual work as well as LLM suggestions. For each sentence in the audio, it will first get interpreted by ASR, then input into LLM model with prompts, and output potential correct answer with reasoning. Then, finally, the transcriptionist would listen to the audio, and choose to pick the suggested correction or not, or add their own correction. The process would now look like the following:

\begin{figure}[h!]
    \centering
    \includegraphics[width=0.8\textwidth, height=0.5\textheight]{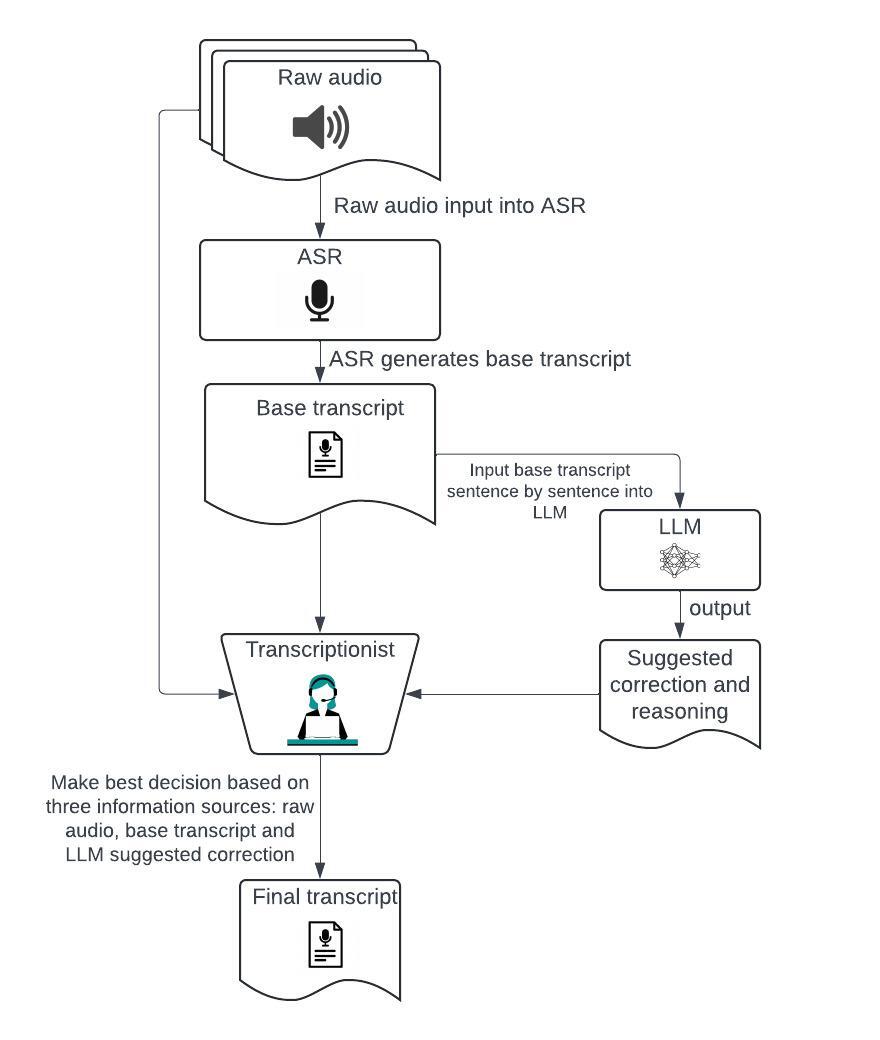}
    \caption{manual+ LLM workflow}
    \label{fig:your-label}
\end{figure}

We experimented this method. With an untrained person as transcriptionist, We provide him the raw audio, base transcript and LLM suggested transcript. Using recording 1 as an example, the final transcript after manually picking correct answers is the following:
\begin{tcolorbox}[breakable,boxrule=0pt, title=Recording 1 Transcript - manual plus LLM correction]
Eighty-year-old female with complex asthma-related shrinking diabetes and some high blood pressure, come under QPDMA, obesity, anxiety, depression, breast cancer in the past, atrial fibrillation on anticoagulation topics abound, heart failure with LVEF around 40\%, hospital with bradycardia and underwent permanent pacemaker placement. Period, she has been doing well since the time without any cardiovascular symptoms. Period, Xeralta has been resumed and patient is back on amino 2 and will be on POO daily. Period, there is no new issues. Period, past medical assurance, no medication is available. All I see is none. Examination temperature, 98.2. Pulse 60 beats per minute. Blood pressure 120 over 80. Saturation, 97\%. General patient allowed to enter if comfortable. Serious, regular, normal. Lungs, no visual crackles. Abdomen soft, no tendons, legs. Trace barotrauma, CNS, no focal neurological risk. Skin normal. Pacemaker site, no bleeding or hematoma. Lab sodium 140, potassium 3.8, BUN 12, creatinine 1.02. TSH is normal. WBC 4, hemoglobin is 13, platelets is 150. Chest x-ray, pacemaker leads are in correct position. Period. Echocardiogram showed LVEF of 55 to 60\%. And mild tricuspid regurgitation. Period, assessment plan. Problem number one, complete heart block. Next in patients with diuretic cardiac does not mean complete heart block. Period, patient was not on any immunoblocking agents. Period, patient has an amiodarone. Period, amiodarone was stopped and patient has taken for permanent pacemaker replacement. Period, patient is doing well. Without any bleeding or hematoma. Period, amiodarone has been restarted. Period, recommended follow-up with electrophysiology team. Period, problem number two, natal fibrillation. Next in patients with paroxysmal natal fibrillation. Period, patient was on a nitrone. Nitrone has been resumed. Period, patient was on apixaban. Nitrone has been resumed. Period, monitor closely for any bleeding or hematoma. Period, problem number three, hypertension, blood pressure is well controlled. Unless no proof of amyloid in a few days. Home blood pressure and heart rate monitoring recommended.
\end{tcolorbox}
\section{Results}
\label{sec:verify}
In this session, we show the WERs (Word Error Rate) and Key Medical Term Error Rate (KMTER) of all the recordings in our dataset, with medical terms we manually picked.
\begin{table}[h]
\caption{Comparison of Results (error rate in percentage)}
\begin{center}
\resizebox{\textwidth}{!}{
\renewcommand{\arraystretch}{1.3}
\begin{tabular}{c|c|c|c|c|c|c|c|c}
\hline
\multicolumn{1}{c|}{\rule{0pt}{12pt}} & \multicolumn{2}{c|}{\textbf{Initial ASR transcript}} & \multicolumn{2}{c|}{\textbf{One Set Result}} & \multicolumn{2}{c|}{\textbf{Sentence by Sentence}} & \multicolumn{2}{c}{\textbf{Manual+LLM}} \\[2pt]
\hline
\textbf{Recording} & \textbf{WERs} & \textbf{KMTER} & \textbf{WERs} & \textbf{KMTER} & \textbf{WERs} & \textbf{KMTER} & \textbf{WERs} & \textbf{KMTER} \\[2pt]
\hline\rule{0pt}{12pt}
Recording 1 & 50.2 & 44.4 & 45.1 & 40.7 & 43.8 & 37.0 & 41.5 & 33.3\\
Recording 2 & 43.5 & 33.9 & 47.8 & 36.0 & 41.7 & 24.5 & 32.1 & 20.9 \\
Recording 3 & 33.2 & 33.9 & 25.4 & 23.1 & 21.2 & 15.8 & 16.2 & 12.5 \\
Recording 4 & 21.5 & 18.9 & 17.8 & 16.7 & 14.3 & 12.8 & 12.1 & 10.5 \\
Recording 5 & 30.8 & 27.5 & 25.7 & 19.3 & 21.2 & 18.3 & 18.5 & 15.3 \\[2pt]
\hline
\end{tabular}
}
\end{center}
\end{table}
From the results, we can see that with the work flow getting fine tuned (from one set input to sentence by sentence), and with manual interact involved, the transcript error rates generally decreases, with some occasional outlier which is understandable, due to the unpredictability of LLM model and the complexity of original audio input. Therefore, we can conclude that, with an appropriate way of prompting LLM models with transcript text from ASR, along with minor manual correction, the overall transcription quality will get largely improved. As a next step of this research, we plan to try out different models and ASRs for our methodology, including some recent models that seem promising. \citep{yuan2024continued}

\section{Conclusion}
\label{sec:conc}
In this research, we proposed an innovative approach to improve medical transcription performance by integrating Large Language Model (LLM)-based Automatic Speech Recognition (ASR) technology with prompt engineering. Our approach leverages the capabilities of LLMs to provide corrections and reasoning, which significantly aids untrained transcriptionists in interpreting accented English and accurately capturing key medical terminologies.

Through a detailed analysis of the corrections and reasoning provided by the LLM, transcriptionists can make more informed decisions, resulting in higher quality transcriptions. By selectively choosing the correct answers between the initial ASR transcript and the LLM corrections, our approach effectively reduces the Word Error Rate (WER) and ensures a higher capture rate of critical medical terms.

Overall, our findings demonstrate that the combination of LLM-based ASR and prompt-engineered LLM models offers a substantial enhancement in the accuracy and reliability of medical transcription. This advancement not only facilitates better patient records but also enhances the overall efficiency and effectiveness of medical documentation processes. Future research could further refine these techniques and explore additional applications within the medical field to continue improving transcription accuracy and reliability.

\paragraph{Notes.}
Experiment and example code can be found here: \url{https://github.com/li001029/Medical_LLM_ASR}

\end{document}